\documentclass[sigconf]{acmart}

\usepackage{algorithm}
\usepackage{algorithmic}
\usepackage{subfigure}
\usepackage{array}
\usepackage{setspace}
\newcolumntype{C}[1]{>{\centering}p{#1}}
\usepackage{float}
\usepackage{url}
\usepackage{balance}
\usepackage{setspace}

\setcopyright{acmcopyright}

\acmYear{2017}
\acmConference{SIGIR'17}{August 07-11, 2017}{Shinjuku, Tokyo, Japan}
\acmISBN{978-1-4503-5022-8/17/08}
\acmPrice{15.00.}
\copyrightyear{2017}
\acmDOI{10.1145/3077136.3080709}

\begin{document}
\title{Top-K Influential Nodes in Social Networks: A Game Perspective}

\author{Yu Zhang}
\affiliation{%
  \institution{Key Laboratory of Machine Perception (MOE) \&}
  \streetaddress{Dept. of Computer Science, Peking University}
  \city{Beijing, China}
}
\email{yuz9@illinois.edu}
\author{Yan Zhang}
\affiliation{%
  \institution{Key Laboratory of Machine Perception (MOE) \&}
  \streetaddress{Dept. of Machine Intelligence, Peking University}
  \city{Beijing, China}
}
\email{zhy@cis.pku.edu.cn}

\begin{spacing}{1.0}
\begin{abstract}
Influence maximization, the fundamental of viral marketing, aims to find top-$K$ seed nodes maximizing influence spread under certain spreading models. In this paper, we study influence maximization from a game perspective. We propose a Coordination Game model, in which every individual makes its decision based on the benefit of coordination with its network neighbors, to study information propagation. Our model serves as the generalization of some existing models, such as Majority Vote model and Linear Threshold model. Under the generalized model, we study the hardness of influence maximization and the approximation guarantee of the greedy algorithm. We also combine several strategies to accelerate the algorithm. Experimental results show that after the acceleration, our algorithm significantly outperforms other heuristics, and it is three orders of magnitude faster than the original greedy method.
\end{abstract}

%
%




\maketitle

\section{Introduction}
Social networks play an important role in information diffusion. They give us the motivation to use a small subset of influential individuals in a social network to activate a large number of people. Kempe et al. \cite{kempekdd03} build a theoretical framework of influence maximization, aiming to find top-$K$ influential nodes under certain spreading models. They discuss two popular models - Independent Cascade (IC) model and Linear Threshold (LT) model and propose a greedy algorithm with $(1-1/e-\epsilon)$-approximation rate.

Easley and Kleinberg \cite{easleykleinberg} divide the cause of information propagation into two categories: information effects and direct-benefit effects. Obviously, IC model and LT model belong to the former one, while we focus on the latter one. In most spreading models, each node has two states: active and inactive. Equivalently saying, it has two choices. In our Coordination Game (CG) model, we regard information diffusion as the process of individual decision-making. As individuals make their decisions based on the benefit of coordination with their network neighbors, a particular pattern of behavior can begin to spread across the links of the network.

Influence maximization under CG model is useful in viral marketing. Let us recall the example in \cite{kempekdd03}. A company would like to market a new product, hoping it will be adopted by a large fraction of the network. The company can initially target a few influential nodes by giving them free samples of the product. Then other nodes will probably switch to using the new product because of the following two reasons: (1) They have a higher evaluation of the new product than the old one. (2) They have to coordinate with their neighbors because using different products may reduce their benefits. (e.g., people using different operating systems may have compatibility problems when working together, and users from different kinds of social media platforms cannot communicate with each other timely.) Our model describes these two reasons precisely.

In this paper, we study how to find Top-$K$ influential nodes under CG model. We first propose our model which serves as the generalization of some well-known spreading models, such as Majority Vote model \cite{chensoda09} and Linear Threshold model \cite{kempekdd03}. We then prove some theoretical results under CG model, including NP-hardness of the optimization problem itself and \#P-hardness of computing the objective function. Then we try to find a good approximation algorithm for the problem. We embed our CG model into the scenario of \textit{general diffusion process} \cite{mossel2010}, and prove that the objective function is monotone and submodular if and only if the \textit{cumulative distribution function} of people's threshold is concave, in which case the greedy algorithm can return a $(1-1/e-\epsilon)$-approximation solution.

As a traditional method, Kempe et al. \cite{kempekdd03} use 10,000 times of Monte Carlo simulations to approximate the objective function, but it costs too much time on large-scale networks. To accelerate our algorithm, we use two efficient heuristics - LazyForward \cite{leskoveckdd07} and StaticGreedy \cite{chengcikm13}. Experimental results show that our \texttt{Greedy} and \texttt{Greedy++} algorithms can activate more nodes than other heuristics. Moreover, \texttt{Greedy++} runs faster than \texttt{Greedy} by three orders of magnitude.

\vspace{1mm}

\noindent \textbf{Related Work.}
Kempe et al. \cite{kempekdd03} first build an algorithmic framework of influence maximization by transforming it into a discrete optimization problem. After their work, a lot of efforts have been made on efficient computing methods of the objective function. Some methods aim to reduce the number of trials that need Monte Carlo simulations, such as \texttt{CELF} \cite{leskoveckdd07}. Other researchers focus on how to calculate the influence spread efficiently. For instance, Chen et al. \cite{chenkdd10, chenicdm10} use arborescences or DAGs to represent the original graph. Cheng et al. propose a \texttt{StaticGreedy} strategy \cite{chengcikm13} and a self-consistent ranking method \cite{chengsigir14}.

Morris \cite{morris} is the first to propose a coordination game model in contagion. This model is also discussed detailedly in Easley and Kleinberg's textbook \cite{easleykleinberg}. We will extend this model by introducing some random factors into utility values.

\section{Model}
In a social network $G=(V,E)$, we study a situation in which each node has a choice between two behaviors, labeled $A$ and $B$. If nodes $u$ and $v$ are linked by an edge, then there is an incentive for them to have their behaviors match. We use a game model to describe this situation. There is a coordination game on each edge $(u,v)\in E$, in which players $u$ and $v$ both have two strategies $A$ and $B$. The payoffs are defined as follows:

\vspace{1mm}

(1) if $u$ and $v$ both adopt strategy $A$, they will get payoffs $p_{uA}>0$ and $p_{vA}>0$ respectively;

(2) if they both adopt strategy $B$, they will get payoffs $p_{uB}>0$ and $p_{vB}>0$ respectively;

(3) if they adopt different strategies, they each get a payoff of 0.

\vspace{1mm}

The payoff matrix is shown in Figure 1.

\begin{figure}
\centering
\includegraphics[scale=0.8]{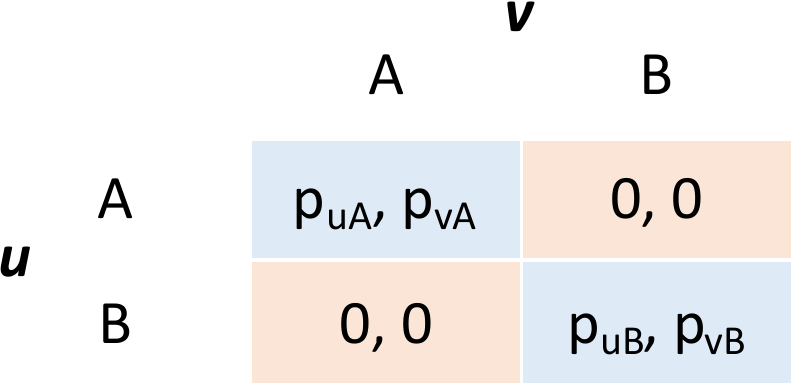}
\caption{Payoff matrix of the coordination game.}
\vspace{-1em}
\end{figure}

We define the total payoff of player $u$ as the sum of the payoffs it gets from all coordination games with its neighbors $N(u) = \{v|(u,v)\in E\}$. If $u$ can get a higher total payoff when it adopts $A$ than that when it adopts $B$, it will choose strategy $A$. Otherwise, it will choose strategy $B$.

According to the actual situation, we have the following assumptions about the payoffs:

\vspace{1mm}

(1) All the $p_{uA}$ and $p_{uB}$ $(u \in V)$ may not be equal to each other because each person in the social network values behaviors $A$ and $B$ differently.

(2) $p_{uA}$ and $p_{uB}$ $(u \in V)$ can either be constants or independent and identically distributed random variables because the cascading behaviors in networks are always considered to have determinate principles with some stochastic factors.

\vspace{1mm}

Suppose $u$ knows all the choices of its neighbors: there are $x_B$ nodes adopting $B$ and $x_A = {\rm deg}(u)-x_B$ nodes adopting $A$. Obviously, $u$ will adopt $B$ if and only if
\begin{equation}
p_{uB}x_B \geq p_{uA}x_A = p_{uA}({\rm deg}(u)-x_B),
\end{equation}
or
\begin{equation}
x_B \geq \frac{p_{uA}}{p_{uA}+p_{uB}}{\rm deg}(u) = \delta_u {\rm deg}(u), \ \ \ \delta_u \in [0,1].
\end{equation}

\vspace{2mm}

\noindent \textbf{Influence Maximization Problem.} Suppose now the market is dominated by $A$ (i.e., all of the nodes in the network choose $A$). Given a constant $k$, we want to find a seed set $S_0 \subseteq V$, $|S_0| \leq k$. Initially, we let each node in $S_0$ adopt $B$ (and they will never change their choices again). Time then runs forward in unit steps. In each step, each node decides whether to switch from strategy $A$ to strategy $B$ according to the payoff-maximization principle. We can regard the evolution of nodes' choices as a spreading process of $B$ in the network. The spread of behavior $B$ will finally stop in at most $n = |V|$ steps.

We define $S_i = |\{u \in V|u$ adopts $B$ in step $i\}|$ $(i=1,2,...,n)$. Our objective function is (the expectation of) the nodes affected by $B$ at last, or
\begin{equation}
\sigma(S_0) = \mathbb{E}_{\{p_{uA},p_{uB}|u \in V\}}[|S_n|] = \mathbb{E}_{\{\delta_u|u \in V\}}[|S_n|].
\end{equation}
Our purpose is to maximize $\sigma(S_0)$ subject to $|S_0| \leq k$.

The CG model can be regarded as the generalization of the following two well-known spreading models.

\vspace{1mm}

\noindent \textbf{Majority Vote Model.} Suppose all the $p_{uA}$ $(u \in V)$ are constants and are equal to each other. So are all the $p_{uB}$ $(u \in V)$. Equivalently, let
\begin{equation}
p_A = p_{uA},\ \ p_B = p_{uB},\ \ \delta = \delta_u = \frac{p_{A}}{p_{A}+p_{B}},\ \ \forall u \in V.
\end{equation}

$\delta$ is a constant threshold same to every nodes. When $p_A = p_B$, or $\delta = \frac{1}{2}$, the spreading model is called Majority Vote model, which is extensively studied in \cite{chensoda09}.

\vspace{1mm}

\noindent \textbf{Linear Threshold Model.} If we set $p_{uA}=1$ and let $p_{uB}$ follow a continuous power-law distribution, i.e., the \textit{probabilistic density function} of $p_{uB}$ is
\begin{equation}
\begin{split}
& f_B(x) = \frac{\alpha}{(x+1)^{\gamma}}\  (x\geq 0), \\
 \text{where}\ \ \gamma>&1 \ \  \text{and} \ \  \alpha = \frac{1}{\int_0^\infty\frac{1}{(x+1)^{\gamma}}{\rm d \it x}} = \gamma - 1,
\end{split}
\end{equation}
then $\forall 0\leq x \leq 1$,
\begin{equation}
\begin{split}
\Pr[\delta_u \leq x] &= \Pr[\frac{1}{1+p_{uB}}\leq x] = \Pr[p_{uB} \geq 1/x-1] \\
&= \int_{1/x-1}^{+\infty}f_B(t){\rm d \it t} = -(t+1)^{-\gamma+1}\bigg|_{1/x-1}^{+\infty} = x^{\gamma-1}.
\end{split}
\end{equation}
If $\gamma = 2$, we will have $\delta_u \sim U[0,1]$. This is the famous Linear Threshold model where the weight on each edge adjacent to node $u$ is $1/{\rm deg}(u)$ (i.e., $b_{vu} = \frac{1}{{\rm deg}(u)}, \forall u,v\in V$).

\vspace{1.5mm}

\noindent \textbf{Hardness.} Under CG model, we have the following hardness result.

\begin{theorem}
\label{thm:hard}
(1) Influence maximization under CG model is NP-hard. (2) Computing the objective function under CG model is $\#$P-hard.
\end{theorem}
The hardness result directly follows the NP-hardness of Influence Maximization under Majority Vote model \cite{chensoda09} and LT model \cite{kempekdd03} and the \#P-hardness of computing the objective function under LT model \cite{chenicdm10}.

\section{Algorithms}
\noindent \textbf{Submodularity.} To find a greedy algorithm with an approximation guarantee,
the submodularity of the objective function is necessary. We first recall the general diffusion process defined by Mossel and Roch in \cite{mossel2010}.

Suppose each node $v$ in the social network $G=(V,E)$ has a threshold $\theta_v \sim U[0,1]$ $i.i.d$ and a ``local" spreading function $f_v:2^{V}\rightarrow [0,1]$. Initially there is a seed set $S_0 \subseteq V$. In each step $t \geq 1$,
\begin{equation}
S_t = S_{t-1} \cup \{v|v\in V-S_{t-1} \ \land \ f_v(S_{t-1})\geq \theta_v\}.
\end{equation}
The spreading process will stop in at most $n = |V|$ steps. So the objective function is $\sigma(S_0) = \mathbb{E}_{\{\theta_u|u \in V\}}[|S_n|]$.

We can embed our model into the scenario of the general diffusion process.

Let $F_{\delta}$ be the cumulative distribution function of $\delta_u$. Since $\delta_u \in [0,1]$, we have $F_{\delta}(0)=0$ and $F_{\delta}(1)=1$. $\forall v$ and $S$, let
\begin{equation}
\theta_v = F_{\delta}(\delta_v)\ \ \text{and} \ \ f_v(S) = F_\delta\Big(\frac{|S \cap N(v)|}{{\rm deg}(v)}\Big).
\end{equation}
Suppose $F_\delta$ is continuous and strictly monotone increasing in $[0,1]$, then $F_\delta^{-1}$ exists, and $\forall x\in [0,1]$,
\begin{equation}
\Pr[F_{\delta}(\delta_v)\leq x] = \Pr[\delta_v \leq F_{\delta}^{-1}(x)] = F_{\delta}(F_{\delta}^{-1}(x)) = x.
\end{equation}
So $F_{\delta}(\delta_v) \sim U[0,1]$.
Therefore
\begin{equation}
\begin{split}
 f_v(S)\geq \theta_v &\Longleftrightarrow F_\delta\Big(\frac{|S \cap N(v)|}{{\rm deg}(v)}\Big)\geq \theta_v \\
 & \Longleftrightarrow |S \cap N(v)|\geq F_\delta^{-1}(\theta_v){\rm deg}(v) \\ & \Longleftrightarrow |S \cap N(v)|\geq \delta_v {\rm deg}(v).
\end{split}
\end{equation}

\begin{lemma}
Suppose $F_\delta$ is continuous and strictly monotone increasing in $[0,1]$, $f_v$ is monotone and submodular for any node $v$ (in any graph)
\textbf{iff} $F_\delta$ is concave in $[0,1]$.
\end{lemma}

It is not difficult to understand Lemma 3.1 intuitively because submodularity can be considered as a kind of concavity. $F_{\delta}$ being concave in $[0,1]$ means that the distribution of people's threshold has a positive skewness, or people tend to have a higher evaluation of new products than old ones. This assumption is reasonable in some cases (e.g., the mobile phone market). $F_\delta$ being continuous and strictly monotone increasing in $[0,1]$ is a technical assumption instead of an essential one. We define these two assumptions as the \textit{concave threshold property}.

For the general diffusion process, Mossel and Roch \cite{mossel2010} have proved that
$\sigma(S_0)$ is monotone and submodular if and only if $f_v$ is monotone and submodular for any $v \in V$.
Therefore, we can get Theorem 3.2 immediately.
\begin{theorem}
$\sigma(S_0)$ is monotone and submodular \textbf{iff} $F_{\delta}$ satisfies the concave threshold property.
\end{theorem}

Theorem 3.2 provides a strong tool to judge the objective function's submodularity under certain spreading models. For example,
under Majority Vote model, $\sigma(S_0)$ is not submodular because $F_\delta(x) = \mathbb{I}(x\geq \delta)$ is not concave in $[0,1]$, where $\mathbb{I}(\cdot)$ is the indicator function. In contrast, under Linear Threshold model, $\sigma(S_0)$ is submodular because $F_\delta(x) = x$ is concave in $[0,1]$.

Up till now, we have proved the monotonicity and submodularity of the objective function under CG model with some necessary assumptions. Using the result in \cite{kempekdd03}, the greedy algorithm given in Algorithm 1 (\texttt{Greedy}) returns a $(1-1/e-\epsilon)$-approximate solution. The algorithm simply selects seed nodes one by one, and each time
it always selects the node that provides the largest marginal gain of the objective function.

\vspace{1mm}

\noindent \textbf{Speeding-Up Algorithm.} Due to the hardness of computing $\sigma(S_0)$, we use two strategies - LazyForward \cite{leskoveckdd07} and StaticGreedy \cite{chengcikm13} to accelerate our algorithm. The reasons why they are useful in submodular cases have been explained in \cite{leskoveckdd07} and \cite{chengcikm13} respectively.

We maintain a priority queue. When finding the next node, we go through the nodes in decreasing order of their marginal gain. If the marginal gain of the top node has not been updated, we recompute it and insert it into the priority queue again.

Instead of conducting a huge number of Monte Carlo simulations each time, we generate a rather small number of snapshots at the very beginning. In all the iterations, we run simulations on these snapshots and use the average to estimate the objective function.

We name the accelerated algorithm as \texttt{Greedy++}.

\begin{algorithm}[htb]
\caption{\texttt{Greedy}($k$, $\sigma$)}
\label{alg:Framwork}
\begin{algorithmic}[1]
\STATE initialize $S_0 = \emptyset$
\FOR {$i = 1$ to $k$}
     \STATE select $u = \arg\max_{v\in V-S_0}(\sigma(S_0\cup\{v\})-\sigma(S_0))$
     \STATE $S_0 = S_0\cup\{u\}$
\ENDFOR
\STATE output $S_0$
\end{algorithmic}
\end{algorithm}

\section{Experiments}
To test the effectiveness and efficiency of our \texttt{Greedy} and \texttt{Greedy++} algorithms, we conduct experiments on three real-world networks and compare our algorithms with other existing heuristics.

\vspace{1mm}

\noindent \textbf{Datasets.} The three real-world datasets include two collaboration networks \texttt{NetHEPT} and \texttt{NetPHY}\footnote{\texttt{http://research.microsoft.com/en-us/people/weic/graphdata.zip}}, and one online social network \texttt{Epinions}\footnote{\texttt{http://snap.stanford.edu/data}}. We summarize the statistical information of the these datasets in Table 1.

\begin{table}[H]
\renewcommand\arraystretch{0.9}
\centering
\caption{Statistical information of three datasets.}
\vspace{-1mm}
\begin{tabular}{|C{1.7cm}|C{1.7cm}|C{1.7cm}|C{1.7cm}|}
\hline
Datasets & $|V|$ & $|E|$ & Type \tabularnewline
\hline
\texttt{NetHEPT} & 15,233 & 58,991 & Undirected \tabularnewline
\texttt{NetPHY} & 37,154 & 231,584 & Undirected \tabularnewline
\texttt{Epinions} & 75,879 & 508,837 & Directed \tabularnewline
\hline
\end{tabular}
\end{table}

\begin{figure*}[!t]
\centering
  \subfigure[Linear Threshold]{
    \label{fig:subfig:a}
    \includegraphics[scale=0.42]{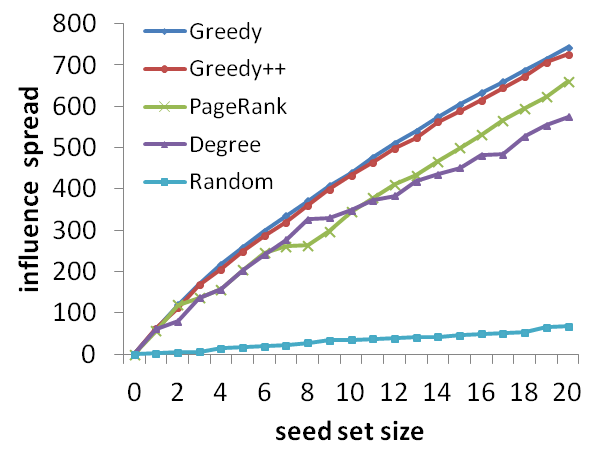}}
  \hspace{-1ex}
  \subfigure[Concave Threshold]{
    \label{fig:subfig:b}
    \includegraphics[scale=0.42]{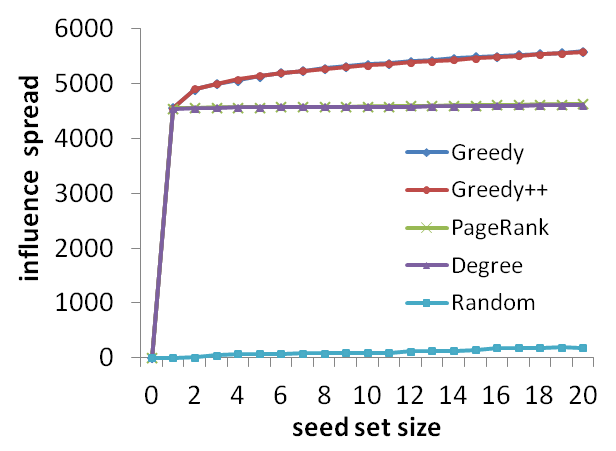}}
  \hspace{-1ex}
  \subfigure[Convex Threshold]{
    \label{fig:subfig:c}
    \includegraphics[scale=0.42]{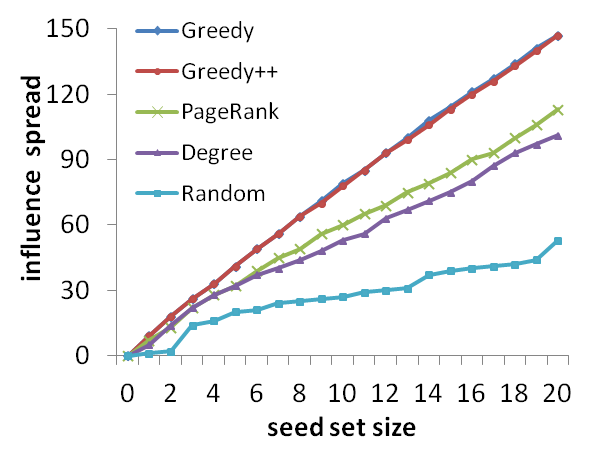}}
  \hspace{-1ex}
  \subfigure[Majority Vote]{
    \label{fig:subfig:d}
    \includegraphics[scale=0.42]{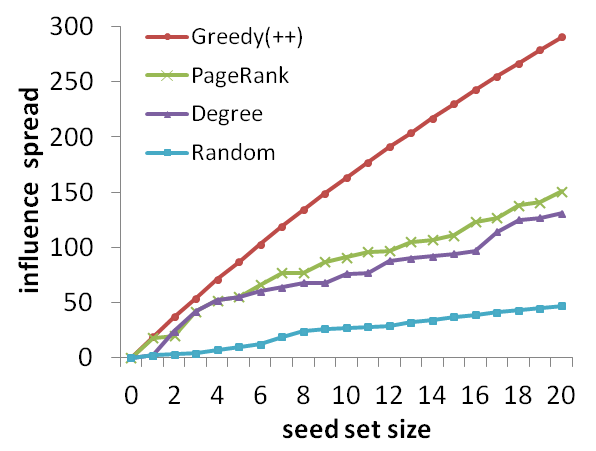}}
\vspace{-0.5em}
  \caption{Influence spread of various algorithms on \texttt{NetHEPT}, with different distribution of $\delta_u$. ($X\sim U[0,1].$) (a) $\delta_u = X$ (submodular). (b) $\delta_u = X^2$ (submodular). (c) $\delta_u = \sqrt{X}$ (nonsubmodular). (d) $\delta_u = 0.5$ (nonsubmodular).}
  \label{fig:subfig}
\vspace{-1em}
\end{figure*}

\noindent \textbf{Algorithms.} A total of five algorithms are tested. Besides \texttt{Greedy} and \texttt{Greedy++} proposed in this paper, we use other three heuristic algorithms as benchmark methods.

\vspace{1mm}

(1) \texttt{PageRank} chooses nodes with the largest PageRank value. For directed networks, influential nodes are considered to have a large number of out-links, while nodes with high PageRank values are considered to have lots of in-links. Therefore, in Epinions, we first change the direction of all edges in the graph and then run PageRank. We use $\alpha = 0.9$ as the random jump parameter.

(2) \texttt{Degree} chooses nodes with the largest out-degree.

(3) \texttt{Random} chooses nodes at random.

There are several other efficient algorithms to solve influence maximization under IC model or LT model, such as \texttt{PMIA} \cite{chenkdd10}, \texttt{LDAG} \cite{chenicdm10} and \texttt{IMM} \cite{tangsigmod15}. However, they cannot be applied in CG model directly, and we will not put them into the comparison.

\vspace{1mm}

\noindent \textbf{Effectiveness.}
We first compare the effectiveness of \texttt{Greedy} and \texttt{Greedy++} with other algorithms by showing influence spread (i.e., $|S_n|$) of the obtained seed set.

In our CG model, the distribution of $\delta_u$ can be various. We run influence maximization algorithms under four different spreading models where $\delta_u$ is $X$, $X^2$, $\sqrt{X}$ and $0.5$, respectively $(X\sim U[0,1])$. Accordingly, the distribution function $F_\delta(x)$ is $x$, $\sqrt{x}$, $x^2$ and $\mathbb{I}(x \geq 0.5)$.

Figure 2 shows our experimental results on \texttt{NetHEPT}. In Figure 2, \texttt{Greedy++} consistently performs on par with \texttt{Greedy} and significantly outperforms other heuristic algorithms in all cases. According to Theorem 3.2, the first two cases are submodular, while the other two are not. However, our experimental results indicate that \texttt{Greedy} and \texttt{Greedy++} still perform well in the non-submodular cases. 
In two larger graphs \texttt{NetPHY} and \texttt{Epinions}, we get similar experimental results.

\vspace{1mm}

\noindent \textbf{Efficiency.}
We now test the running time of these algorithms. Figure 3 shows our experimental results.

\begin{figure}
\centering
  \subfigure[]{
    \label{fig:subfig:a}
    \includegraphics[scale=0.42]{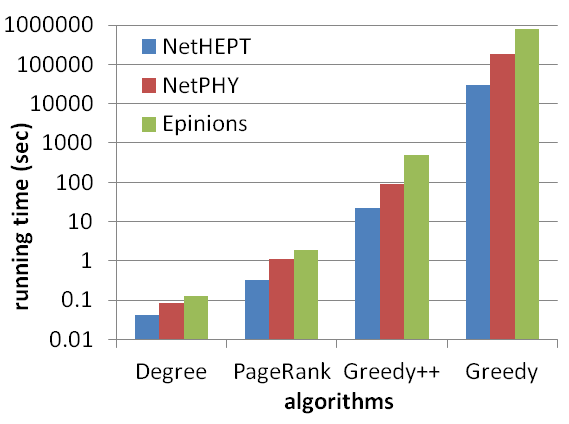}}
  \hspace{-1ex}
  \subfigure[]{
    \label{fig:subfig:b}
    \includegraphics[scale=0.42]{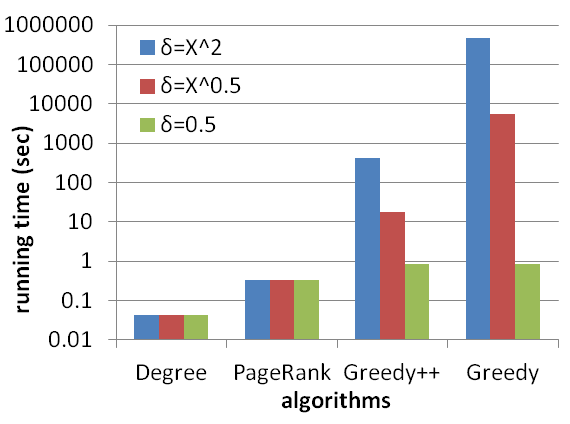}}
\vspace{-1em}
  \caption{(a) Running time of various algorithms on three datasets with $F_\delta(x) = x$.\ \ 
           (b) Running time of various algorithms on \texttt{NetHEPT} with different distributions of $\delta_u$\ ($X\sim U[0,1].$)}
  \label{fig:subfig}
\vspace{-1.5em}
\end{figure}

As we expected, \texttt{Greedy++} runs consistently faster than \texttt{Greedy}, with more than three orders of magnitude speedup. For example, in the linear threshold case, it takes \texttt{Greedy} more than 9 days to get the top-20 influential nodes on \texttt{Epinions}, while \texttt{Greedy++} only requires about 8 minutes.

In the concave threshold case, \texttt{Greedy++} spends more time because $\delta_u$ is small and the influence spread tends to be wide. However, this is worthwhile because the strategies only finding ``central nodes" no longer work in this case (see Figure 2(b)).

\section{Conclusions}
In this paper, we have discussed how to find top-$K$ influential nodes in social networks under a game theoretic model. We show the hardness of the optimization problem itself, as well as the hardness of calculating the objective function. We prove the approximation guarantee of the greedy algorithm under necessary assumptions. We also accelerate our algorithm with the combination of LazyForward and StaticGreedy. Our experimental results demonstrate that \texttt{Greedy++} matches \texttt{Greedy} in the spreading effect while significantly reduces running time, and it outperforms other heuristic algorithms such as MaxDegree and PageRank.

\vspace{1mm}

\noindent\textbf{Acknowledgements.} This work is supported by 973 Program under Grant No.2014CB340405, NSFC under Grant No.61532001 and  No.61370054. We thank the anonymous reviewers for their valuable comments.

\bibliographystyle{ACM-Reference-Format}
{
\small
\bibliography{sigir17}


\begin{thebibliography}{00}


\ifx \showCODEN    \undefined \def \showCODEN     #1{\unskip}     \fi
\ifx \showDOI      \undefined \def \showDOI       #1{#1}\fi
\ifx \showISBNx    \undefined \def \showISBNx     #1{\unskip}     \fi
\ifx \showISBNxiii \undefined \def \showISBNxiii  #1{\unskip}     \fi
\ifx \showISSN     \undefined \def \showISSN      #1{\unskip}     \fi
\ifx \showLCCN     \undefined \def \showLCCN      #1{\unskip}     \fi
\ifx \shownote     \undefined \def \shownote      #1{#1}          \fi
\ifx \showarticletitle \undefined \def \showarticletitle #1{#1}   \fi
\ifx \showURL      \undefined \def \showURL       {\relax}        \fi
\providecommand\bibfield[2]{#2}
\providecommand\bibinfo[2]{#2}
\providecommand\natexlab[1]{#1}
\providecommand\showeprint[2][]{arXiv:#2}

\bibitem[\protect\citeauthoryear{Chen}{Chen}{2009}]%
        {chensoda09}
\bibfield{author}{\bibinfo{person}{N. Chen}.} \bibinfo{year}{2009}\natexlab{}.
\newblock \showarticletitle{On the Approximability of Influence in Social
  Networks}. In \bibinfo{booktitle}{{\em SODA'09}}. \bibinfo{publisher}{SIAM},
  \bibinfo{address}{Austin, Texas, USA}, \bibinfo{pages}{1029--1037}.
\newblock


\bibitem[\protect\citeauthoryear{Chen, Wang, and Wang}{Chen
  et~al\mbox{.}}{2010a}]%
        {chenkdd10}
\bibfield{author}{\bibinfo{person}{W. Chen}, \bibinfo{person}{C. Wang}, {and}
  \bibinfo{person}{Y. Wang}.} \bibinfo{year}{2010}\natexlab{a}.
\newblock \showarticletitle{Scalable influence maximization for prevalent viral
  marketing in large-scale social networks}. In \bibinfo{booktitle}{{\em
  KDD'10}}. \bibinfo{publisher}{ACM}, \bibinfo{address}{Washington, DC, USA},
  \bibinfo{pages}{1029--1038}.
\newblock


\bibitem[\protect\citeauthoryear{Chen, Yuan, and Zhang}{Chen
  et~al\mbox{.}}{2010b}]%
        {chenicdm10}
\bibfield{author}{\bibinfo{person}{W. Chen}, \bibinfo{person}{Y. Yuan}, {and}
  \bibinfo{person}{L. Zhang}.} \bibinfo{year}{2010}\natexlab{b}.
\newblock \showarticletitle{Scalable influence maximization in social networks
  under the linear threshold model}. In \bibinfo{booktitle}{{\em ICDM'10}}.
  \bibinfo{publisher}{IEEE}, \bibinfo{address}{Sydney, Australia},
  \bibinfo{pages}{88--97}.
\newblock


\bibitem[\protect\citeauthoryear{Cheng, Shen, Huang, Chen, and Cheng}{Cheng
  et~al\mbox{.}}{2014}]%
        {chengsigir14}
\bibfield{author}{\bibinfo{person}{S. Cheng}, \bibinfo{person}{H. Shen},
  \bibinfo{person}{J. Huang}, \bibinfo{person}{W. Chen}, {and}
  \bibinfo{person}{X. Cheng}.} \bibinfo{year}{2014}\natexlab{}.
\newblock \showarticletitle{Imrank: Influence maximization via finding
  self-consistent ranking}. In \bibinfo{booktitle}{{\em SIGIR'14}}.
  \bibinfo{publisher}{ACM}, \bibinfo{address}{Gold Coast, Australia},
  \bibinfo{pages}{475--484}.
\newblock


\bibitem[\protect\citeauthoryear{Cheng, Shen, Huang, Zhang, and Cheng}{Cheng
  et~al\mbox{.}}{2013}]%
        {chengcikm13}
\bibfield{author}{\bibinfo{person}{S. Cheng}, \bibinfo{person}{H. Shen},
  \bibinfo{person}{J. Huang}, \bibinfo{person}{G. Zhang}, {and}
  \bibinfo{person}{X. Cheng}.} \bibinfo{year}{2013}\natexlab{}.
\newblock \showarticletitle{Staticgreedy: solving the scalability-accuracy
  dilemma in influence maximization}. In \bibinfo{booktitle}{{\em CIKM'13}}.
  \bibinfo{publisher}{ACM}, \bibinfo{address}{San Francisco, CA, USA},
  \bibinfo{pages}{509--518}.
\newblock


\bibitem[\protect\citeauthoryear{Easley and Kleinberg}{Easley and
  Kleinberg}{2010}]%
        {easleykleinberg}
\bibfield{author}{\bibinfo{person}{D. Easley} {and} \bibinfo{person}{J.
  Kleinberg}.} \bibinfo{year}{2010}\natexlab{}.
\newblock \bibinfo{booktitle}{{\em Networks, crowds, and markets: Reasoning
  about a highly connected world}}.
\newblock Cambridge University Press.
\newblock


\bibitem[\protect\citeauthoryear{Kempe, Kleinberg, and Tardos}{Kempe
  et~al\mbox{.}}{2003}]%
        {kempekdd03}
\bibfield{author}{\bibinfo{person}{D. Kempe}, \bibinfo{person}{J. Kleinberg},
  {and} \bibinfo{person}{{\'E}. Tardos}.} \bibinfo{year}{2003}\natexlab{}.
\newblock \showarticletitle{Maximizing the spread of influence through a social
  network}. In \bibinfo{booktitle}{{\em KDD'03}}. \bibinfo{publisher}{ACM},
  \bibinfo{address}{Washington, DC, USA}, \bibinfo{pages}{137--146}.
\newblock


\bibitem[\protect\citeauthoryear{Leskovec, Krause, Guestrin, Faloutsos,
  VanBriesen, and Glance}{Leskovec et~al\mbox{.}}{2007}]%
        {leskoveckdd07}
\bibfield{author}{\bibinfo{person}{J. Leskovec}, \bibinfo{person}{A. Krause},
  \bibinfo{person}{C. Guestrin}, \bibinfo{person}{C. Faloutsos},
  \bibinfo{person}{J. VanBriesen}, {and} \bibinfo{person}{N. Glance}.}
  \bibinfo{year}{2007}\natexlab{}.
\newblock \showarticletitle{Cost-effective outbreak detection in networks}. In
  \bibinfo{booktitle}{{\em KDD'07}}. \bibinfo{publisher}{ACM},
  \bibinfo{address}{San Jose, CA, USA}, \bibinfo{pages}{420--429}.
\newblock


\bibitem[\protect\citeauthoryear{Morris}{Morris}{2000}]%
        {morris}
\bibfield{author}{\bibinfo{person}{S. Morris}.}
  \bibinfo{year}{2000}\natexlab{}.
\newblock \showarticletitle{Contagion}.
\newblock \bibinfo{journal}{{\em The Review of Economic Studies\/}}
  \bibinfo{volume}{67} (\bibinfo{year}{2000}), \bibinfo{pages}{57--78}.
\newblock


\bibitem[\protect\citeauthoryear{Mossel and Roch}{Mossel and Roch}{2010}]%
        {mossel2010}
\bibfield{author}{\bibinfo{person}{E. Mossel} {and} \bibinfo{person}{S. Roch}.}
  \bibinfo{year}{2010}\natexlab{}.
\newblock \showarticletitle{Submodularity of influence in social networks: From
  local to global}.
\newblock \bibinfo{journal}{{\it SIAM J. Comput.}} \bibinfo{volume}{39},
  \bibinfo{number}{6} (\bibinfo{year}{2010}), \bibinfo{pages}{2176--2188}.
\newblock


\bibitem[\protect\citeauthoryear{Tang, Shi, and Xiao}{Tang
  et~al\mbox{.}}{2015}]%
        {tangsigmod15}
\bibfield{author}{\bibinfo{person}{Y. Tang}, \bibinfo{person}{Y. Shi}, {and}
  \bibinfo{person}{X. Xiao}.} \bibinfo{year}{2015}\natexlab{}.
\newblock \showarticletitle{Influence maximization in near-linear time: a
  martingale approach}. In \bibinfo{booktitle}{{\em SIGMOD'15}}.
  \bibinfo{publisher}{ACM}, \bibinfo{address}{Melbourne, Australia},
  \bibinfo{pages}{1539--1554}.
\newblock


\bibitem[\protect\citeauthoryear{Valiant}{Valiant}{1979}]%
        {valiant}
\bibfield{author}{\bibinfo{person}{L.~G. Valiant}.}
  \bibinfo{year}{1979}\natexlab{}.
\newblock \showarticletitle{The complexity of enumeration and reliability
  problems}.
\newblock \bibinfo{journal}{{\it SIAM J. Comput.}}  \bibinfo{volume}{8}
  (\bibinfo{year}{1979}), \bibinfo{pages}{410--421}.
\newblock


\end{thebibliography}
}
\end{spacing}

\appendix
\section{Proof of Theorem 2.1}
\begin{proof}
(1) Chen \cite{chensoda09} proves the NP-hardness of Influence Maximization under Majority Vote model with $\delta = \frac{1}{2}$, which is enough to demonstrate the first result.

(2) Chen et al. \cite{chenicdm10} prove it is $\#$P-hard to compute exact influence in general networks under LT model. They use the settings that
$b_{vu} = const, \forall u,v\in V$ in their proof.
We modify the proof and get the hardness result under our settings.\footnote{Note that $b_{vu} = \text{const}$ is not a special case of CG model.} We reduce this problem from the problem of counting simple paths in a directed graph. Given a directed graph $G=(V,E)$, counting the total number of simple paths in $G$ is $\#$P-hard \cite{valiant}. Let $n=|V|$ and $D=\max_{v\in V}{\rm deg}_{in}(v)$. From $G$, we construct $n+1$ graphs $G_1,G_2,$ $...,G_{n+1}$. To get $G_i$ $(1\leq i \leq n+1)$, we first add $D+i-{\rm deg}_{in}(v)$ ``branching nodes" linking to node $v$ for all $v \in V$. And then we add a node $s$ linking to all nodes in $V$. Thus each node in $G_i$ has $D+i+1$ in-links except ``branching nodes" and $s$.

According to our assumption, the weight on each edge in $G_i$ is $w_i = \frac{1}{D+i+1}$. Let $S_0 = \{s\}$ and $\mathcal{P}$ denote the set of all simple paths starting from $s$ in $G_i$. (Note that $\mathcal{P}$ is identical in all $G_i$ because ``branching nodes" are unreachable from $s$.) According to \cite{chenicdm10}, we have
\begin{equation}
\sigma_{G_i}(S_0) = \sum_{\pi\in\mathcal{P}}\prod_{e\in\pi}w_i,\ \ \ (1\leq i\leq n+1),
\end{equation}
where $\sigma_{G_i}(S_0)$ means $\sigma(S_0)$ in $G_i$. Let $B_j$ be the set of simple paths of length $j$ in $\mathcal{P}$ $(0\leq j \leq n)$. We have
\begin{equation}
\sigma_{G_i}(S_0) = \sum_{j=0}^n\sum_{\pi\in B_j}\prod_{e\in\pi}w_i = \sum_{j=0}^n\sum_{\pi\in B_j}w_i^{j} = \sum_{j=0}^n w_i^{j}|B_j|.
\end{equation}

We want to solve these $n+1$ linear equations with $n+1$ variables $|B_0|,|B_1|,...,|B_n|$. Since the coefficient matrix is a Vandermonde matrix, $(|B_0|,|B_1|,...,|B_n|)$ is unique and easy to compute.

Finally, we notice that for each $j = 1,2,...,n$, there is a one-to-one correspondence between paths in $B_j$ and simple paths of length $j-1$ in $G$. Therefore, $\sum_{j=1}^n |B_j|$ is the total number of simple paths in $G$. We complete our reduction.
\end{proof}

\section{Proof of Lemma 3.1}
\begin{proof}
($\Leftarrow$) If $F_\delta$ is concave in $[0,1]$, let $g_v(S) = \frac{|S\cap N(v)|}{{\rm deg}(v)}$, which is a modular function.
It is easy to prove that the composition of a concave function and a modular function is submodular. Therefore $f_v = F_\delta \circ g_v$ is also monotone and submodular.

($\Rightarrow$) If $F_\delta$ is not concave in $[0,1]$, then $\exists a,b,\lambda \in [0,1]$ such that
\begin{equation}
\lambda F_\delta(a) + (1-\lambda)F_\delta(b) > F_\delta(\lambda a + (1-\lambda)b).
\end{equation}
Since $F_\delta$ is (uniformly) continuous and bounded, if we pick up three rational numbers $\frac{N_1}{M}, \frac{N_2}{M}$ and $\frac{p}{q}$ which
are very close to $a,b,\lambda$ respectively, we will have
\begin{equation}
\frac{p}{q} F_\delta\Big(\frac{N_1}{M}\Big) + \frac{q-p}{q}F_\delta\Big(\frac{N_2}{M}\Big) > F_\delta\Big(\frac{N_1p+N_2(q-p)}{Mq}\Big) = F_\delta\Big(\frac{N_3}{Mq}\Big).
\end{equation}

Let $X_i = (\frac{i}{Mq},F_\delta(\frac{i}{Mq}))$ be the points on the curve of $F_\delta$ $(i = N_1q, ... ,N_2q)$ and $l_0$ be the line across $X_{N_1q}$ and $X_{N_2q}$. We know that $X_{N_3}$ is below $l_0$. Therefore $\exists K_1 \leq N_3-1$ and $K_2 \geq N_3$ such that

(1) $X_{K_1}$ is above or in $l_0$ while $X_{K_1+1}$ is below $l_0$.

(2) $X_{K_2}$ is below $l_0$ while $X_{K_2+1}$ is above or in $l_0$.

Let $l_1$ be the line across $X_{K_1}$ and $X_{K_1+1}$ and let $l_2$ be the line across $X_{K_2}$ and $X_{K_2+1}$. We know
that $k(l_1) < k(l_0) < k(l_2)$, where $k()$ is the slope of the line.

Assume there is a node $v$ with $Mq$ neighbors. Let $S$ be the set of $v$'s $K_1$ neighbors and $T$ be the set of $v$'s $K_2$ neighbors,
where $S \subset T$. There is another neighbor $u \notin T$. Therefore
\begin{equation}
\begin{split}
&f_v(T\cup \{u\}) - f_v(T) = F_\delta\Big(\frac{K_2+1}{Mq}\Big) - F_\delta\Big(\frac{K_2}{Mq}\Big) = \frac{k(l_2)}{Mq} \\
&>\frac{k(l_1)}{Mq} = F_\delta\Big(\frac{K_1+1}{Mq}\Big) - F_\delta\Big(\frac{K_1}{Mq}\Big) = f_v(S\cup \{u\}) - f_v(S),
\end{split}
\end{equation}
which violates the submodularity of $f_v$.
\end{proof}

\section{The \texttt{Greedy++} Algorithm}
\begin{algorithm}[H]
\caption{\texttt{Greedy++}($k$, $\sigma$, $R'$)}
\label{alg:greedypp}
\small
\begin{algorithmic}[1]
\STATE initialize $S_0 = \emptyset$
\FOR {$i = 1$ to $R'$}
     \STATE generate the threshold $\delta_v$ $(\forall v \in V)$ for snapshot $G_i$
\ENDFOR
\FORALL {$v \in V$}
     \STATE $\Delta_v = +\infty$ //initialize the marginal gain of each node
\ENDFOR
\FOR {$i = 1$ to $k$}
     \FORALL {$v \in V-S_0$}
         \STATE $cur_v = $ False
     \ENDFOR
     \WHILE {True}
         \STATE $u = \arg\max_{v \in V-S_0}\Delta_v$ //maintain a priority queue
         \IF {$cur_u$}
             \STATE $S_0 = S_0 \cup \{u\}$
             \STATE \textbf{break}
         \ELSE
             \STATE $\Delta_u = \frac{1}{R'}\sum_{i=1}^{R'}(\sigma_{G_i}(S_0\cup\{u\})-\sigma_{G_i}(S_0))$
             \STATE reinsert $u$ into the priority queue and heapify
             \STATE $cur_u = $ True
         \ENDIF
     \ENDWHILE
\ENDFOR
\STATE output $S_0$
\end{algorithmic}
\end{algorithm}

\section{Additional Experimental Results}
\begin{figure}[H]
\centering
  \subfigure[\texttt{NetPHY}]{
    \includegraphics[scale=0.39]{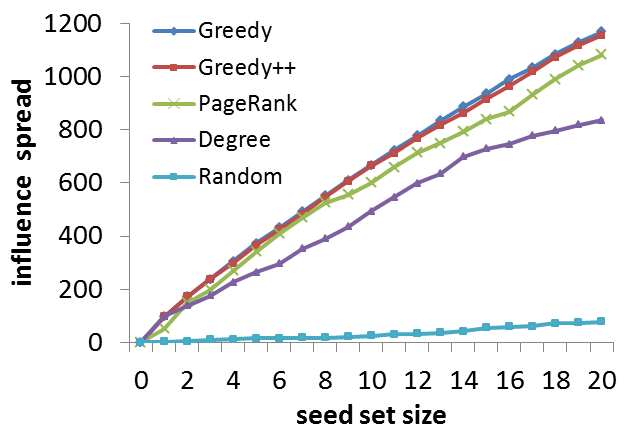}}
  \hspace{-1.5ex}
  \subfigure[\texttt{Epinions}]{
    \includegraphics[scale=0.39]{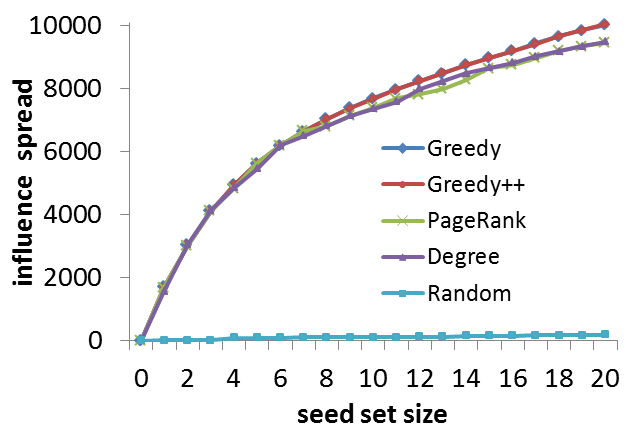}}
  \vspace{-1em}
  \caption{Influence spread of various algorithms on (a) \texttt{NetPHY} and (b) \texttt{Epinions}. ($F_\delta(x) = x$.)}
\end{figure}

\end{document}